# Concept based Attention

Jie You[1], Xin Yang[1], Matthias Hub[2]

[1] Dept. of Electronics and Information Engineering, Huazhong University of Sci. & Tech., China

[2] Dept. of Computer Science, University of Tubingen, Germany

**Abstract:** Attention endows animals an ability to concentrate on the most relevant information among a deluge of distractors at any given time, either through volitionally 'top-down' biasing, or driven by automatically 'bottom-up' saliency of stimuli, in favour of advantageous competition in neural modulations for information processing. Nevertheless, instead of being limited to perceive simple features, human and other advanced animals adaptively learn the world into categories and abstract concepts from experiences, imparting the world meanings. This thesis suggests that the high-level cognitive ability of human is more likely driven by attention basing on abstract perceptions, which is defined as concept based attention (CbA).

Attention endows animals an ability to concentrate on the most relevant information among a deluge of distractors at any given time, either through volitionally 'top-down' biasing, or driven by automatically 'bottom-up' saliency of stimuli, in favour of advantageous competition in neural modulations for information processing (Desimone & Duncan, 1995; Knudsen, 2007; Itti & Koch, 2001; Kastner & ungerleider, 2000). Literally, many studies about visual attention have examined the effects and neural mechanisms of shifting attention between different locations in the visual field (Reynolds & Chelazzi, 2004), and among different properties of diverse feature dimensions, such as orientation, color, speed or direction of motion (Maunsell & Treue, 2006). Meanwhile, some biologically-inspired computational models are addressed for object recognition, by integrating location-based and feature-based biasing signals along the message-passing of Bayesian Network (Rao, 2005).

Nevertheless, instead of being limited to perceive simple features, human and other advanced animals adaptively learn the world into categories and abstract concepts from experiences, imparting the world meanings. With this high-level cognitive ability, we usually pay attention to or are attracted by some abstract concepts, such as "neuroscience" and "beautiful girl", rather than specific features. Intuitively, when readers are reading this paragraph they may be paying attention on the concept: concept-based attention (CbA). Someone may argue that the concepts can be decomposed into or described by a set of causally related features, and that when attention is directed to high-level concepts it is actually feature-based. However, on one hand it is conceivably difficult to uniquely point out what features a concept include in a specific situation, because different individual may treat it in different ways, depending on their experience and knowledge, and besides, for one individual a concept can vary with time (Fig. 1).

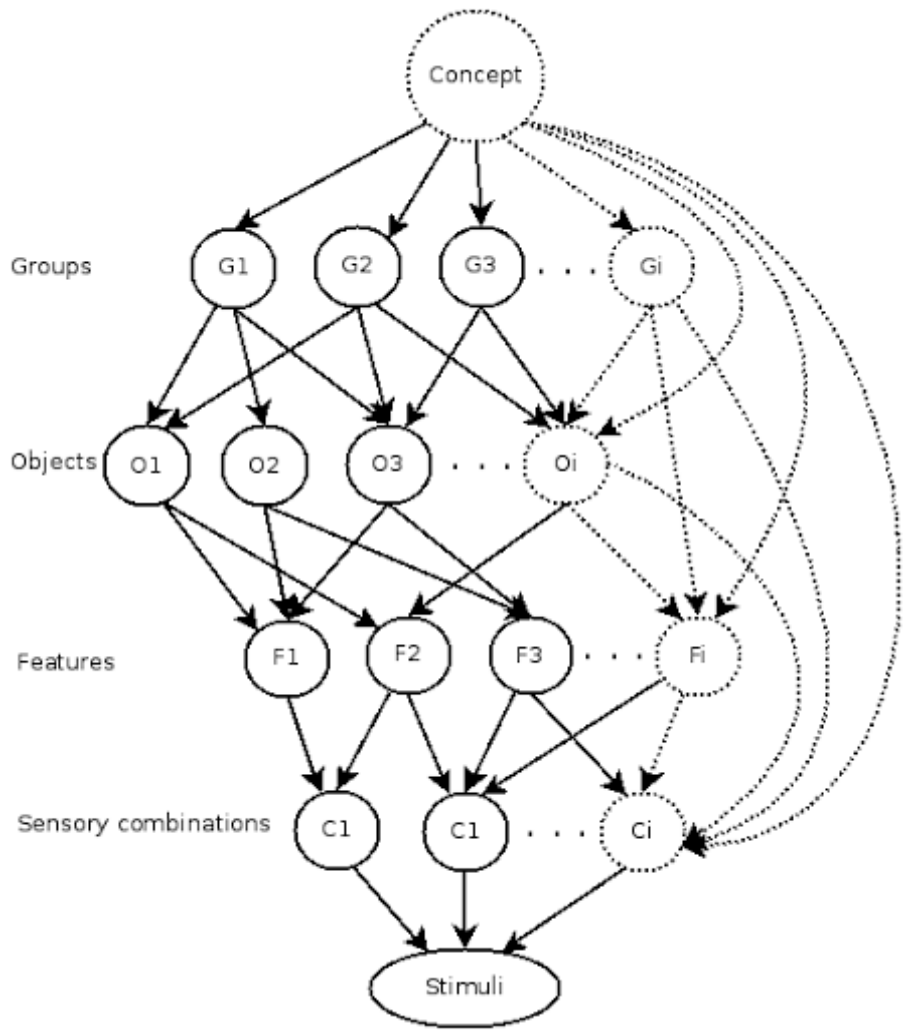

Fig 1: A schematic diagram of information relation and interaction represented in the neural network. The circles indicate a node in Bayesian Network and the arrows indicate the dependent relationship between nodes. The probabilistic relationships of the nodes are very arbitrary, largely depending on subject's learning history and experience. Therefore, a category or concept may include many different features and a feature may be components of many categories.

On the other hand, It is possible that representing the behavioural relevance of all the features would require more neurons than the brain could afford (Maunsell & Treue, 2006), and to implement a topographic map for the representation of behaviourally relevant features in large dimensions would be inefficient or even impossible for the neural network. Furthermore, from object recognition point of view, whereas identification is easier to implement in artificial vision system, categorization seems to be the simpler and more immediate stage in recognition process in biological visual systems (Riesenhuber & Poggio, 2000; Ullman, 1996; Logothetis & Sheinberg, 1996). It is thereby reasonable to hypothesize that neurons prefer attend to high-level abstracts over detailed specialties, unless they are particularly required. Finally, since evidences from neuropsychological and neuroimaging studies in mammalian brains has revealed communications between anatomically distant neural regions through oscillatory synchronizations underlying cognitive behavior (Buschman & Miller, 2007; Womelsdorf et al., 2007; Saalmann et al., 2007), we should have reason to believe that these neural regions preferentially choose to encode information in more efficient way when communicating to each other just as we communicate in language, neurons or neuronal coalitions convey preferences with some form of high-level symbols representing categories and concepts, and indeed they have the ability to encode the abstracts (Seger & Miller, 2010; Wallis et al., 2001; Freedman et al., 2001; Miller, 2003), because more efficient communication mediates more competitive information for accessing working memory or even recalling long-term memory.

When we are thinking about something interesting (subjectively interested)|a game, a mathematical formula or a business plan while listening a speech report, we can hear nothing, see nothing even if staring at the animate pictures shown by the speaker, which nevertheless engages other listener's attention, because the information processing capacity of our brain is completely seized by the attended thing (manipulation of concepts). CbA seems to spontaneously engage and sustain a bootstrapping recurrent loop at the higher-level of information processing hierarchy in the brain. Moreover, consistent with the location- and feature-based attention, CbA largely depends on the information about animal's internal state (Knudsen, 2007), but could be disconnected with external world at a point of time, self-engaging in working

memory. Therefore, examining the neural correlates of CbA would help understand why and how people think in a particular way, and might shed light on exploring the psychology of interest (Silvia, 2006). Further, CbA can integrate information across multiple sensory modalities, biasing information processing at a higher level. For example, attractive look, fragrant smell and savoury taste collectively dene 'delicious food', attracting gourmet's interest. Thus, it could also facilitate understanding how the interaction of structured knowledge and statistic inference mediate inductive learning and reasoning (Tenenbaum et al., 2006). Last, from the point of view regarding attention deficit, besides direct connection with the degraded function of PFC for working memory it may be contributed by the inability of categorization, that is, due to lack of abstracting information into meaningful categories (efficient compression of information) the brain may be maddened by welter of details, which in turn cause hyperactivity. Study on CbA might therefore help unravel aspects of the pathology of attention deficit hyperactivity disorder (ADHD).

As the neural mechanisms of category learning have been widely investigated in primate prefrontal cortex (Seger & Miller, 2010; Freedman et al., 2001). I plan to examine the neural correlates of CbA rst from studying the neurophysiological mechanisms and ring patterns representing and directing categorical information for attentional modulation in PFC and relevant cortical areas. Based on the experiments reported on the references (Buschman & Miller, 2007; Freedman et al., 2001), an electrophysiological experiment is designed as: two monkeys perform DMS task for visual search, and the visual target and distractors are from two different categories (according to arbitrary rules, such as visual similarity, depending on how we train the monkeys). Physiological data are recorded in PFC and lateral intraparietal (LIP) areas across the trials before the monkeys are trained for categorization and after, including reaction time, ring rate, spiking frequency and coherence. After training for categorization, the visual search task trials are divided into three kinds: first, the visual target on the sample view and matching view is the same one object; second, target on the former and the later view are different but from the same category; last, visual target on the sample view is a symbol, such as a 'D' for dog (monkeys have been trained to recognize it), indicating the category, and the visual target on the matching view is an instance of the category (see Fig. 2).

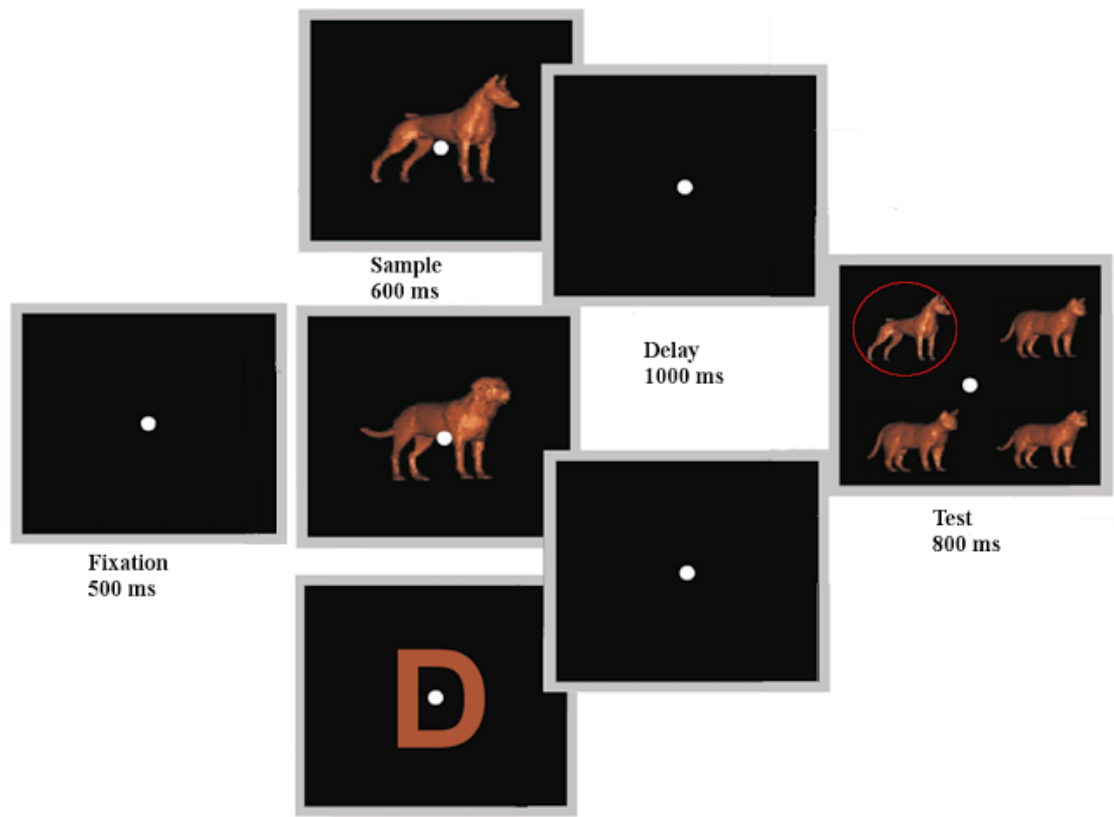

Fig 2: Behavioural task for visual search. Visual stimuli are generated from three-dimensional morphing system for two categories: 'cats' and 'dogs' (Freedman et al., 2001). White filled-circle indicates the fixation point and the red circle indicates the visual target.

Two-way ANOVA (planned and unplanned) will be used for the data analysis in large dimensions would be inefficient or even impossible for the neural network. Furthermore, from object recognition point of view, whereas identification is easier to implement in artificial vision system, categorization seems to be the simpler and more immediate stage in recognition process in biological visual systems (Riesenhuber & Poggio, 2000; Ullman, 1996; Logothetis & Sheinberg, 1996). It is thereby reasonable to hypothesize that neurons prefer attend to high-level abstracts over detailed specialties, unless they are particularly required. Finally, since evidences from neuropsychological and neuroimaging studies in mammalian brains has revealed communications between anatomically distant neural regions through oscillatory synchronizations underlying cognitive behavior (Buschman & Miller, 2007; Womelsdorf et al., 2007; Saalmann et al., 2007), we should have reason to believe that these neural regions preferentially choose to encode information in more efficient way when communicating to each other just as we communicate in language, neurons or neuronal coalitions convey preferences with some form of high-level symbols representing categories and concepts, and indeed they have the ability to encode the abstracts (Seger & Miller, 2010; Wallis et al., 2001; Freedman et al., 2001; Miller, 2003), because more efficient communication mediates more competitive information for accessing working memory or even recalling long-term memory.